\documentclass{article} 
\usepackage{iclr2027_conference,times}

\usepackage{amsmath,amsfonts,bm}

\def\eqref#1{equation~\ref{#1}}

\def\1{\bm{1}}

\DeclareMathAlphabet{\mathsfit}{\encodingdefault}{\sfdefault}{m}{sl}
\SetMathAlphabet{\mathsfit}{bold}{\encodingdefault}{\sfdefault}{bx}{n}

\usepackage{url}
\usepackage{xcolor}
\definecolor{mycite}{HTML}{0072BC}
\usepackage[colorlinks=true, citecolor=mycite, linkcolor=mycite, urlcolor=mycite]{hyperref}
\usepackage{url}            
\usepackage{booktabs}       
\usepackage{amsfonts}       
\usepackage{microtype}      
\usepackage{xcolor}         
\usepackage{amsmath}
\usepackage{algorithm}
\usepackage[most]{tcolorbox}
\usepackage{enumitem}
\usepackage{multirow}
\usepackage[dvipsnames]{xcolor}
\usepackage{pifont}   
\usepackage{amsthm}
\usepackage{algpseudocode}  
\usepackage{wrapfig}
\usepackage{tabularx}
\usepackage{amssymb}
\usepackage[table]{xcolor}
\usepackage{makecell}
\usepackage{kotex}
\usepackage[dvipsnames]{xcolor}

\newcommand{\red}[1]{{\color{red}#1}}

\newtheorem{observation}{Observation}

\usepackage{cleveref}
\crefname{appendix}{App.}{Apps.}
\Crefname{appendix}{App.}{Apps.}
\crefname{section}{Sec.}{Secs.}
\Crefname{section}{Sec.}{Secs.}
\crefname{equation}{Eq.}{Eqs.}
\Crefname{equation}{Eq.}{Eqs.}
\crefname{figure}{Fig.}{Figs.}
\Crefname{figure}{Fig.}{Figs.}
\crefname{table}{Tab.}{Tabs.}
\Crefname{table}{Tab.}{Tabs.}
\crefname{theorem}{Theorem}{Theorems}
\Crefname{theorem}{Theorem}{Theorems}
\crefname{proposition}{Proposition}{Propositions}
\Crefname{proposition}{Proposition}{Propositions}
\crefname{lemma}{Lemma}{Lemmas}
\Crefname{lemma}{Lemma}{Lemmas}
\crefname{remark}{Remark}{Remarks}
\Crefname{remark}{Remark}{Remarks}
\crefname{definition}{Definition}{Definitions}
\Crefname{definition}{Definition}{Definitions}
\crefname{observation}{Observation}{Observations}
\Crefname{observation}{Observation}{Observations}

\usepackage[most]{tcolorbox}
\usepackage{xcolor}
\definecolor{lavender}{RGB}{240, 240, 250}   
\definecolor{mint}{RGB}{229,242,229}  

\definecolor{gray}{RGB}{246,246,246}  
\definecolor{cfblue0}{HTML}{E8EFFF}  
\definecolor{cfblue1}{HTML}{e0eaff}  
\definecolor{cfblue2}{HTML}{c4d7ff}  
\definecolor{cfblue3}{HTML}{a4c1ff}  
\definecolor{cfblue4}{HTML}{7fa8ff}  
\definecolor{cfblue5}{HTML}{7499ff}  
\definecolor{beige}{HTML}{FAF8F3}  
\definecolor{cfblue6}{HTML}{1866B1}  
\definecolor{cfblue7}{HTML}{3067FF}
\newcommand{\cfblue}[1]{{\color{cfblue7}#1}}

\newtcolorbox{plainbox_cfblue}{
    enhanced,
    breakable,
    colback=cfblue0,
    colframe=cfblue0,
    boxrule=0pt,
    left=3pt,
    right=3pt,
    top=5pt,
    bottom=5pt,
    before skip=5pt,
    after skip=5pt
}

\newtcolorbox{plainbox_mint}{
    enhanced,
    breakable,
    colback=mint,
    colframe=mint,
    boxrule=0pt,
    left=3pt,
    right=3pt,
    top=5pt,
    bottom=5pt,
    before skip=5pt,
    after skip=5pt
}

\newtcolorbox{plainbox_lavender}{
    enhanced,
    breakable,
    colback=lavender,
    colframe=lavender,
    boxrule=0pt,
    left=3pt,
    right=3pt,
    top=5pt,
    bottom=5pt,
    before skip=5pt,
    after skip=5pt
}

\newtcolorbox{plainbox_gray}{
    enhanced,
    breakable,
    colback=gray,
    colframe=gray,
    boxrule=0pt,
    left=10pt,
    right=10pt,
    top=2pt,
    bottom=2pt,
    before skip=3pt,
    after skip=10pt
}

\newtcolorbox{plainbox_beige}{
    enhanced,
    breakable,
    colback=beige,
    colframe=beige,
    boxrule=0pt,
    left=3pt,
    right=3pt,
    top=5pt,
    bottom=5pt,
    before skip=5pt,
    after skip=5pt
}

\definecolor{algbg}{HTML}{FAF8F3} 
\definecolor{algcomment}{HTML}{005E0B} 
\definecolor{algkw}{HTML}{4B3F8C}
\definecolor{algfunc}{HTML}{2E75B6}
\definecolor{algemph}{HTML}{9B2C2C}

\lstdefinestyle{idf}{
  language=Python,
  backgroundcolor=\color{algbg},
  basicstyle=\ttfamily\scriptsize,
  commentstyle=\color{algcomment},
  keywordstyle=\color{algkw}\bfseries,
  emph={randn,rerope,range,len,append},
  emphstyle=\color{algfunc},
  emph={[2]G,E},
  emphstyle={[2]\color{algemph}\bfseries},
  showstringspaces=false,
  columns=fullflexible,
  xleftmargin=8pt,
  framexleftmargin=8pt,
  aboveskip=4pt, belowskip=4pt,
}
\usepackage{relsize}
\newcommand{\emailfont}[1]{{\mdseries\relscale{0.95}\texttt{#1}}}

\usepackage{listings}

\title{In-Distribution Forcing\\for Long Video Generation at Test Time}

\author{%
\textbf{Jeongwoo Shin}$^{1*}$, \quad
\textbf{Youngyoon Choi}$^{1*}$, \quad
\textbf{Sangwoo Jo}$^{2}$, \quad
\textbf{Hyunmog Kim}$^{2}$, \\
\textbf{Sungjoon Choi}$^{2\dagger}$, \quad
\textbf{Joonseok Lee}$^{1\dagger}$, \quad
\textbf{Jaewoong Choi}$^{3\dagger}$, \quad
\textbf{Jaemoo Choi}$^{4\dagger}$ \\
{\mdseries $^{1}$Seoul National University, \ $^{2}$Korea University,} \\
{\mdseries $^{3}$Sungkyunkwan University, \ $^{4}$Georgia Institute of Technology} \\
\emailfont{\{swswss, youngyoon911, joonseok\}@snu.ac.kr} \\
\emailfont{\{jasonjo97, sungjoonc\}@korea.ac.kr, hyunmog2020@alumni.kaist.ac.kr}\\
\emailfont{jaewoongchoi@skku.edu, jchoi843@gatech.edu}
}

\iclrfinalcopy 
\begin{document}

\maketitle
{\renewcommand{\thefootnote}{}%
 \footnotetext{$^{*}$Equal contribution. \quad $^{\dagger}$Corresponding author.}}
 
\vspace{-.8cm}
\begin{figure}[H]
    \centering
    \includegraphics[width=1.\linewidth]{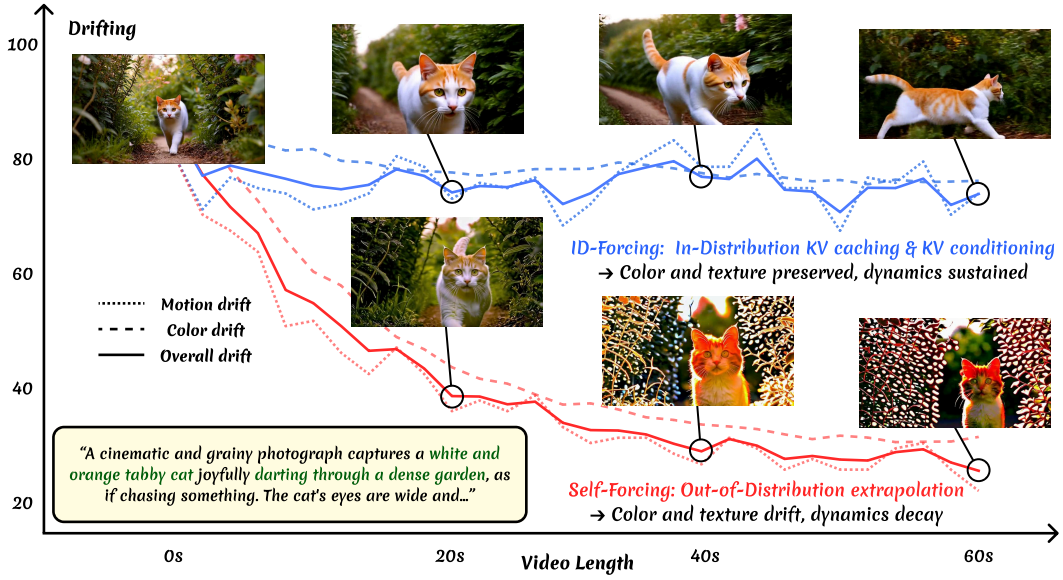}
    \vspace{-.4cm}
    \caption{
        \textbf{ID-Forcing extends autoregressive video generation far beyond the training horizon.} In-distribution KV operations suppress color and motion drift, enabling stable minute-scale generation.
    }
    \label{fig:intro}
    \vspace{-.4cm}
\end{figure}
\begin{abstract}
Modern autoregressive (AR) video diffusion models excel at short-horizon video generation, yet generating long videos remains challenging due to \emph{drifting}, where colors and textures shift, and motion dynamics decay. Existing works primarily rely on \emph{KV conditioning}, which selects or modifies cached key-value (KV) entries to mitigate drifting. However, we observe that KV conditioning alone is insufficient as it assumes cached KV entries remain in-distribution. This assumption fails beyond the training horizon: nothing constrains the construction of KV entries during rollout, giving rise to the \emph{KV-provenance problem} where cached entries themselves become out-of-distribution (OOD). To address this, we propose \textbf{In-Distribution Forcing (ID-Forcing)}, a test-time framework that aligns both KV caching and KV conditioning with training configurations. 
Its key mechanism, \emph{self-caching}, prevents OOD KV entries at their source. Each chunk is cached without attending to prior KV entry, keeping the rolling window exactly in-distribution.
Consequently, ID-Forcing seamlessly extends short-horizon models to minute-scale video generation. Extensive evaluations show that our method remains competitive on standard video generation benchmark while substantially outperforming prior work in mitigating drifting, as validated by both our drift metrics and a user study. 
\par
{\centering \textbf{Project page:} \url{https://in-distribution-forcing.github.io/}\par}
\end{abstract}
\section{Introduction}
\label{sec:intro}

Recent advances in video diffusion models~\citep{veo3,klingomni,sora2,wan} enable high-fidelity synthesis of short video clips with coherent motion, 
yet interactive applications such as world models~\citep{relic,lyra2,voyager}, streaming content~\citep{streamdit,streamingt2v,skyreelsv2}, and long-form storytelling~\citep{infinitystory,shotstream} demand generation far beyond a few seconds. To enable such long-horizon generation, autoregressive (AR) video diffusion~\citep{causvid,diffusionforcing,selfforcing,longlive} has emerged as the standard paradigm, generating videos sequentially in chunks while conditioning each new chunk on the key--value (KV) cache of previously generated content. Specifically, Self-Forcing~\citep{selfforcing} reduces the training-inference gap by
training on autoregressive rollouts, while LongLive~\citep{longlive} extends generation to longer horizons through streaming long-video tuning.



However, the ability to generate over extended horizons does not ensure that visual quality persists. As generation proceeds, colors and textures tend to shift, and motion dynamics gradually decay. This progressive degradation, known as \cfblue{\emph{drifting}}, arises because the rollout at inference differs from the rollout seen during training.
Training on self-rollouts~\citep{selfforcing,selfforcingpp,longlive,contextforcing,rollingforcing} addresses this by exposing the model to its own generated context, but only within a finite training horizon. Extrapolating beyond this horizon therefore introduces conditioning configurations outside the training distribution, causing errors to accumulate over time. Yet simply extending the training horizon cannot solve this issue, as extrapolating past the horizon is inherently unavoidable in arbitrarily long video generation.

To address drifting, a large body of work controls what the model reads from the KV cache during generation, which we refer to as \cfblue{\emph{KV conditioning}}.
Existing approaches typically select cached entries at the chunk level, as with attention sinks~\citep{deepforcing}, or modify their temporal positional embeddings~\citep{infinityrope,memrope}.
These methods, however, are valid only under the assumption that the cached KV entries  are in-distribution, as they consider only which entries to attend to and where to position them.
This assumption has not been explicitly examined, and nothing ensures that it holds.
Thus, understanding drifting requires examining not only the KV entries read during generation, but also the context under which they were constructed.



\begin{figure}[t]
    \centering
    \includegraphics[width=1.\linewidth]{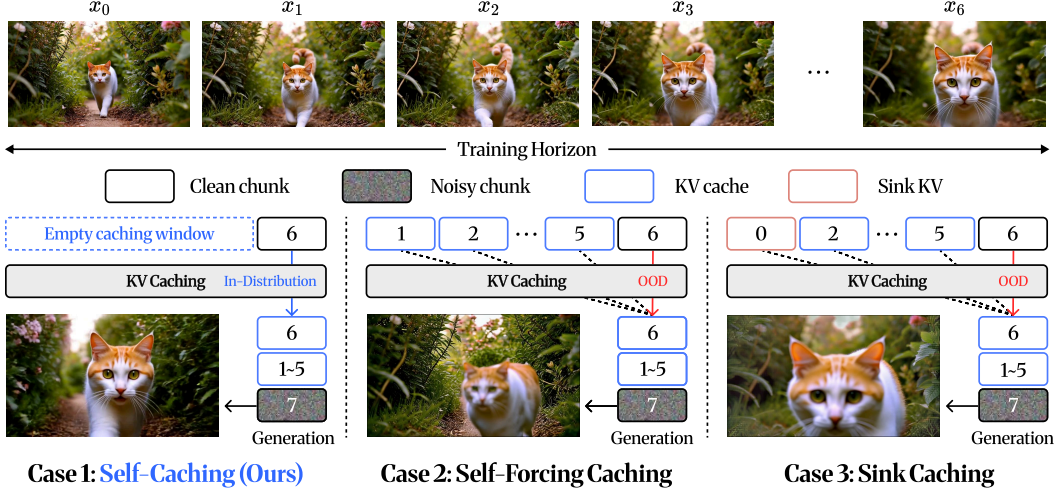}
    \vspace{-.5cm}
    \caption{
        \textbf{Same conditioning, different provenance.} 
        We generate the next chunk $x_7$ from the same conditioning window $\kappa_{1:6}$, varying only how the last chunk of the horizon, $x_6$, is cached into $\kappa_6$.
        \textbf{Case 1} (self-caching): $x_6$ attends to no prior entry, and $x_7$ shows no drifting.
        \textbf{Case 2} (self-forcing caching): $x_6$ attends to a window never seen in training, and $x_7$ degrades.
        \textbf{Case 3} (sink caching): keeping the first chunk does not help, as the window is still unseen in training.
        Thus, how a chunk is cached alone can determine whether the next chunk drifts.
    }
    \label{fig:kv_provenance}
    \vspace{-.2cm}
\end{figure}

We observe that whether a KV entry is in-distribution is determined by \cfblue{\emph{KV caching}}, the operation that computes and stores a KV entry for each generated chunk.
A KV entry depends not only on its own chunk but also on the other KV entries it attends to during KV caching, \emph{i.e.,} its \cfblue{\emph{provenance}}.
Beyond the training horizon, every chunk is cached under a provenance never seen during training, so the resulting KV entry is out-of-distribution.
We term this the \cfblue{\emph{KV-provenance problem}} (\cref{fig:kv_provenance}).
Since provenance is fixed at KV caching, no KV conditioning policy can recover from it, and keeping extrapolation in-distribution requires controlling \emph{both KV caching and KV conditioning}.

Motivated by this insight, we propose \textbf{In-Distribution Forcing (ID-Forcing)}, a test-time method that aligns both KV caching and KV conditioning with the KV context configurations encountered during training. 
For KV caching, we introduce \cfblue{\emph{self-caching}}.
Beyond the training horizon, the earlier KV entries are cached attending only to themselves, while the later ones are cached autoregressively on top of them, which prevents any unseen provenance from being constructed (\S\ref{sec:provenance}).
For KV conditioning, we show that the first chunk must remain in the conditioning window for it to stay in-distribution, and that self-caching keeps the rolling window exactly in-distribution (\S\ref{sec:conditioning}).
Every KV operation is thus one that training executed.
As a result, our method yields a remarkably simple framework:
a plain rolling window with one rule for each KV operation, extending a five-second model to minute-scale generation.

Our contributions are as follows:
\begin{itemize}[leftmargin=2em, itemsep=1pt, topsep=1pt]
    \item We characterize the \textbf{KV-provenance problem} as a source of drift in long-video extrapolation and establish the importance of keeping both KV caching and KV conditioning within the configurations encountered during training (\S\ref{sec:provenance}).
    \item We propose \textbf{In-Distribution Forcing (ID-Forcing)}, which introduces \textbf{self-caching} to control cache provenance and jointly aligns KV caching and KV conditioning with training (\S\ref{sec:provenance}, \S\ref{sec:conditioning}). 
    \item We demonstrate effective suppression of drift in minute-scale generation, preserving visual quality and temporal consistency while maintaining active motion (\S\ref{sec:exp}).
\end{itemize}
\vspace{-.3cm}

\section{Preliminary}
\label{sec:prelim}
\textbf{Autoregressive Video Diffusion.}
A latent video is a sequence of chunks $x_{j} \in \mathbb{R}^{f\times h\times w\times d}$, each covering $f$ latent frames, generated under a text condition $c$.
Chunked autoregressive (AR) diffusion models~\citep{causvid,diffusionforcing,selfforcing} implement the factorization
\begin{align}
    p\big(x_{0:M}\,\big|\,c\big)
    \;=\;
    \prod_{{j}=0}^{M} \, p\big(x_{j} \,\big|\, x_{<{j}},\, c\big).
    \label{eq:factorization}
\end{align}
The conditioning on $x_{<{j}}$ is not implemented on raw latents: each chunk is written into a single key--value (KV) entry once generated, and later chunks are conditioned on these KV entries through the attention operation. We omit the shared condition $c$ below.

\textbf{KV Operations during Training.}
AR factorization in \cref{eq:factorization} is implemented by performing self-rollouts over a fixed horizon of $N$ chunks ($i = 0, \dots, N-1$) using a few-step denoiser $G_\theta$ with $T$ denoising steps~\citep{selfforcing}.
Here, we write $x_i^t$ for chunk $i$ at noise level $t \in \{0,\dots,T\}$, where $x_i^{T}$ is pure noise and $x_i^{0}=x_i$ is clean.
In this process, each step relies on two core KV operations: \emph{KV conditioning} for $x_i$ generation and \emph{KV caching} for storing the KV states of $x_i$ to generate the next chunk.

\cfblue{\emph{KV conditioning.}}
To generate chunk $x_i$, the denoiser $G_\theta$ performs $T$ denoising steps while attending to all preceding KV entries $\kappa_{<i}$:
\begin{align}
    x_i^{t-1} \;=\; G_\theta\!\big(x_i^{t},\, t\,;\; \kappa_{<i}\big),
    \qquad t = T,\dots,1,
    \quad x_i^{T}\!\sim\!\mathcal{N}(0,I).
    \label{eq:gen}
\end{align}

\cfblue{\emph{KV caching.}}
After generating the clean chunk $x_i$, the same denoiser at $t=0$ computes its corresponding KV entry $\kappa_i$ by attending to the same KV set $\kappa_{<i}$:
\begin{align}
    \kappa_i
    \;=\; E\big(x_i^{0};\, \kappa_{<i}\big)
    \;=\; \mathrm{KV}\Big[\,\overline{G}_\theta\big(x_i^{0},\, t{=}0\,;\; \kappa_{<i}\big)\Big].
    \label{eq:cache}
\end{align}
Here, $\mathrm{KV}[\cdot]$ extracts the KV states from the forward pass while discarding the denoising prediction, and the bar denotes a stop-gradient operator. For $i=0$, the initial chunk attends solely to itself. As shown in \cref{eq:gen,eq:cache}, the conditioning window and the caching window are the same set $\kappa_{<i}$, which grows from $0$ to $N{-}1$ during training.

\textbf{Drift during Extrapolation.}
Beyond the training horizon $i\ge N$, standard rolling-window methods fix both the window sizes to $L< N$, evicting the oldest entry $\kappa_{i{-}L}$ when generating each chunk. 
KV conditioning (\cref{eq:gen}) and KV caching (\cref{eq:cache}) are therefore unchanged in form, but both now attend to the $L$ most recent KV entries $\kappa_{i{-}L:i{-}1}$ rather than the full prefix $\kappa_{<i}$. 
This manifests as quality degradation known as \emph{drifting} (exposure bias)~\citep{diffusionforcing,selfforcing}: colors and textures shift, and motion dynamics decay.

\textbf{Temporal Positions.}
Attention uses rotary position embeddings (RoPE~\citep{rope}); we consider only the temporal axis in this work.
Rotations are applied to queries and keys, and attention logits depend only on the relative temporal gap between them, not on their absolute indices.
A chunk occupies a \emph{slot} in the attention window, and each slot receives the $f$ consecutive temporal indices.
We write $R_\Delta$ for re-rotation of a KV entry's stored keys by $\Delta$ frames.

\section{In-Distribution Forcing}
\begin{figure}[t]
    \centering
    \includegraphics[width=1.\linewidth]{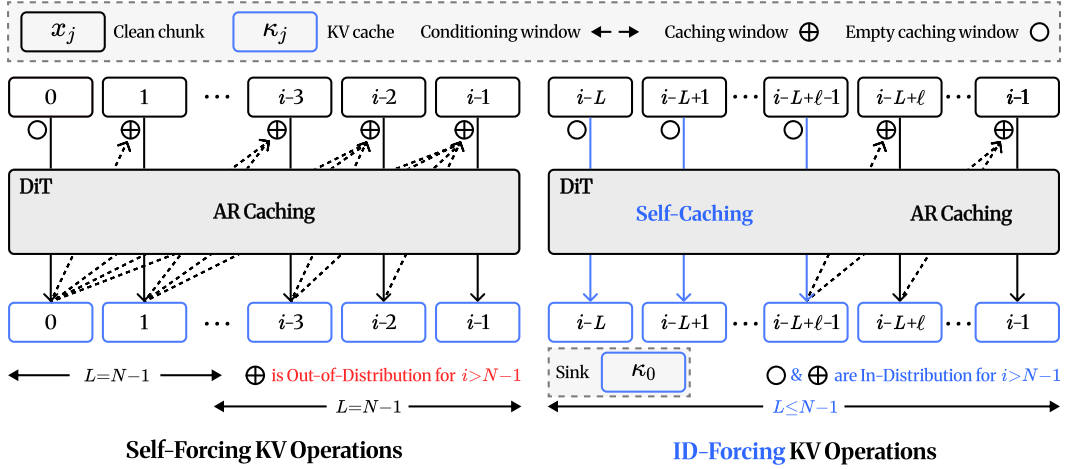}
    \vspace{-.6cm}
    \caption{
        \textbf{KV operations comparison.} Both illustrations show the caching window of each KV entry and the conditioning window of size $L$ used to generate chunk $x_i$. \textbf{Left}: Self-Forcing caches every entry autoregressively, so entries beyond the horizon are cached under windows unseen in training. \textbf{Right}: ID-Forcing caches the earlier entries with self-caching and the rest autoregressively on top of them, so every entry is cached under a window seen in training.
    }
    \label{fig:kvcaching_comparison}
    \vspace{-.2cm}
\end{figure}

\label{sec:method}
We propose \textbf{In-Distribution Forcing (ID-Forcing)}, a training-free framework that keeps test-time extrapolation within the training distribution.
Prior works address drifting by modifying KV conditioning, treating drifting as a consequence of discarding informative past context~\citep{deepforcing,infinityrope,memrope}. 
We instead view drifting as a deviation from the training distribution, and show that this deviation occurs not just in KV conditioning, but also in KV caching.
ID-Forcing therefore introduces two-level control over KV operations: KV caching and KV conditioning.

\begin{plainbox_beige}
\begin{itemize}[label={}, itemsep=3pt, topsep=0pt, parsep=0pt, partopsep=0pt, leftmargin=*]
    \item \textbf{Level 1 (KV caching):} How each KV entry is written (\S\ref{sec:provenance}).
    \item \textbf{Level 2 (KV conditioning):} How KV entries are read during chunk generation (\S\ref{sec:conditioning}).
\end{itemize}
\end{plainbox_beige}


\subsection{Level 1 -- KV Caching: Enforcing In-Distribution Provenance}
\label{sec:provenance}

Our key observation is that the same clean chunk $x_i$ maps to different points in KV space depending on its \cfblue{\emph{provenance}}, the existing KV entries within the caching window (\Cref{fig:kv_provenance}).
Crucially, when this provenance involves unseen context, the resulting KV entry $\kappa_i$ falls out-of-distribution (OOD), and we term this the \cfblue{\emph{KV-provenance problem}}.

Beyond the horizon, a rolling window introduces the KV-provenance problem. Subsequently, every later generation that reads the resulting KV entries inherits their OOD state, which accumulates as drifting.
To resolve this problem, we propose \textbf{Self-Caching}, which keeps KV caching strictly within the training distribution throughout the entire extrapolation.

\textbf{Motivation.} We first inspect the generation process (\cref{eq:gen}) of an arbitrary beyond-horizon chunk $x_i$.
At its first denoising step, $x_i$ is initialized from pure noise, \emph{i.e.}, $x_i^{T}\!\sim\!\mathcal{N}(0,I)$, and its queries, keys, and values are generated via frozen projections.
Consequently, the input itself remains in-distribution, leaving the KV conditioning window as the only possible entry point for drift.
\begin{plainbox_gray}
\begin{observation}
\label{obs:first_ood_gen}
For every $i\ge N$, the KV entries in the conditioning window are the sole path through which drift can enter the generation of chunk $x_i$.
\end{observation}
\end{plainbox_gray}


The question is then whether those KV entries are themselves in-distribution. This depends on the caching window each of them attended to when cached, \emph{i.e.}, its provenance.

\begin{plainbox_gray}
\begin{observation}
\label{obs:firstood}
The KV entries read when generating chunk $x_i$ are in-distribution only if each of them was cached under a provenance that training executed.
\end{observation}
\end{plainbox_gray}

The two observations show that generating $x_i$ in-distribution requires in-distribution KV entries in the conditioning window. These entries, in turn, are in-distribution only if their provenance is.

\cfblue{\textbf{Self-Caching.}}
We propose a simple caching rule that keeps the provenance of each KV entry in-distribution throughout the extrapolation.
Concretely, we restrict every provenance to the configurations that training executed, so that no KV entry is ever written under an unseen context.


For a conditioning window of $L$ KV entries, we cache the first $\ell$ entries attending only to themselves, $E(x_j;\varnothing)$.
The remaining $L{-}\ell$ entries are then cached autoregressively on top of the last self-cached entry, mirroring how training caches $\kappa_1,\kappa_2,\dots$ on top of $\kappa_0$.
\cref{fig:kvcaching_comparison} visualizes the proposed method.

\begin{plainbox_beige}
\textbf{KV caching rule.} 
For every $i\ge N$, the conditioning window of length $L< N$ consists of the KV entries $\kappa_{i{-}L},\dots,\kappa_{i{-}1}$, which are cached as:
\begin{align}
\kappa_j \;=\;
\begin{cases}
E\big(x_j;\,\varnothing\big), & i{-}L \le j < i{-}L{+}\ell,\\[2pt]
E\big(x_j;\,\kappa_{i{-}L{+}\ell{-}1\,:\,j-1}\big), & i{-}L{+}\ell \le j \le i{-}1 .
\end{cases}
\label{eq:writerule}
\end{align}
\end{plainbox_beige}
Starting from the training horizon, every chunk is generated from in-distribution KV entries and is itself cached as an in-distribution one, so merely depending on earlier chunks does not cause drift.
\subsection{Level 2 -- KV Conditioning: Enabling Exact Rolling Window}
\label{sec:conditioning}
Self-caching keeps every stored KV entry in-distribution, but how these entries are read must also match training. We address KV conditioning from two perspectives: (1) the distributional role of the first chunk, and (2) the exact rolling window enabled by self-caching.

\begin{wrapfigure}{r}{0.48\textwidth}
\vspace{-1.1em}
\refstepcounter{algorithm}\label{alg:idforcing}
\hrule height 1pt
\vspace{2pt}
{\textbf{Algorithm \thealgorithm} In-Distribution Forcing}
\vspace{2pt}
\hrule height 0.6pt
\begin{lstlisting}[style=idf]
# G: T-step denoiser, E: KV caching
# N, M: training, target horizon
# L: window size, n_self: self-cached

X, KV = horizon_chunks(), self_cached_KV()

for i in range(N, M):
    # sink + L-1 recent, re-rotated
    W = [KV[0]] + rerope(KV[i-L+1:i])

    # re-cache the tail autoregressively
    for j in range(n_self, L):
        W[j] = E(X[i-L+j], W[n_self-1:j])

    # KV conditioning
    x = randn(*shape)
    for t in range(T, 0, -1):
        x = G(x, t, W)

    # Self-caching
    X.append(x)
    KV.append(E(x, []))

return X
\end{lstlisting}
\vspace{2pt}
\hrule height 1pt
\vspace{-1.2em}
\end{wrapfigure}





\textbf{First Chunk as a Persistent Condition.} 
During training, every generation is conditioned on $\kappa_0$, meaning the conditioning window always contains the first chunk within the training horizon. 
However, as the window rolls during extrapolation, $\kappa_0$ is evicted; yet 
$\kappa_0$ is in fact itself a necessary condition for staying in-distribution at extrapolation.

Unlike later chunks, the causal 3D-VAE~\citep{wan} encodes the initial latent frame from a single pixel frame, whereas every subsequent frame aggregates multiple frames.
$\kappa_0$ is therefore distributionally irreplaceable, and evicting it once the window rolls ($i \ge N$) is itself out-of-distribution.
This differs from an attention sink~\citep{attentionsink,deepforcing} in both motivation and form.
Attention sinks retain several early KV entries for their empirical effect of stabilizing generation, whereas we retain exactly the first one because it alone is distributionally unique.



\textbf{Bounded Window and Exact Rolling.}
We constrain our conditioning window to a length of $L< N$, holding $\kappa_0$ at slot $0$ and the $L{-}1$ most recent KV entries in the remaining slots. This strictly bounds the total window size within the training budget $N{-}1$. After generating each chunk, the oldest non-sink KV entry is evicted and the remaining ones are re-rotated by $R_{-f}$, while $\kappa_0$ remains fixed at slot 0. Consequently, every temporal distance within the window matches a state observed during training.

Self-caching is what keeps this rolling window exactly in-distribution.
During training, every cached KV entry coexists with its complete provenance set because the training window only appends entries without eviction. Under test-time rolling, however, eviction removes past context, leaving the remaining KV entries without their original provenance. By contrast, a self-cached KV entry possesses no external provenance to lose; thus, eviction leaves its representation strictly within the training distribution. 
The complete procedure of our method is summarized in \Cref{alg:idforcing}.

\begin{plainbox_beige}
\textbf{KV conditioning rule.} 
For every $i\ge N$, the oldest KV entry other than $\kappa_0$ is evicted and the remaining KV entries are re-rotated by $R_{-f}$ before chunk $i$ reads them:
\begin{align}
\{\kappa_0\} \cup \{\kappa_{i-L+1},\dots,\kappa_{i-1}\}
\;\leftarrow\;
\{\kappa_0\} \cup R_{-f}\big[\{\kappa_{i-L+1},\dots,\kappa_{i-1}\}\big] .
\label{eq:readrule}
\end{align}
\end{plainbox_beige}

\section{Experiments}
\label{sec:exp}

We evaluate ID-Forcing on minute-scale video generation, extrapolating far beyond the training horizon on two base models, Self-Forcing~\citep{selfforcing} and LongLive~\citep{longlive}.
In \S\ref{sec:main_results}, we compare ID-Forcing with existing test-time KV conditioning methods on 120- and 240-second generation through quantitative metrics, qualitative comparisons, and a user study.
In \S\ref{sec:ablation}, we examine how each component of ID-Forcing; self-caching for KV caching, and the first-chunk sink and re-rotation for KV conditioning, contributes to suppressing drifting.

\subsection{Experimental Setup}
\textbf{Implementation Details.} We implement our method on top of the Wan2.1-T2V-1.3B~\citep{wan} architecture and evaluate it across two
base models: Self-Forcing~\citep{selfforcing} and LongLive~\citep{longlive}. 
We compare against prior test-time extrapolation methods for autoregressive video diffusion, Deep Forcing~\citep{deepforcing}, $\infty$-RoPE~\citep{infinityrope}, and MemRoPE~\citep{memrope}.
Each video chunk consists of 3 latent frames, generated using a 4-step denoising schedule at timesteps $\{1000, 750, 500, 250\}$. We use conditioning window length $L = 3$ and self-caching length $\ell = 2$ for Self-Forcing. For LongLive, which is trained on minute-long videos, we apply our method from the first chunk with $L = 3$ and $\ell = 3$. In both cases, $\ell$ counts the sink $\kappa_0$.


\textbf{Evaluation.} We generate 120- and 240-second videos at a resolution of $480 \times 832$ using the prompts from MovieGen~\citep{moviegen}. Following previous works~\citep{causvid,selfforcingpp,deepforcing,memrope}, we use the first 128 prompts which are refined using Qwen2.5-7B-Instruct. We report metrics from the VBench-Long~\citep{vbench} benchmark, along with drift-specific metrics: color drift~\citep{ttc} and motion drift. Details for these metrics are provided in App.~\S\ref{app:impl_detail}. To further validate automated metrics, we conduct a user study following the 2AFC protocol~\citep{deepforcing, memrope}.


\definecolor{firstbg}{HTML}{a4c1ff}  
\definecolor{secondbg}{HTML}{e0eaff} 
\definecolor{groupbg}{HTML}{F5F5F5}   

\newcommand{\first}[1]{\cellcolor{firstbg}\textbf{#1}}
\newcommand{\second}[1]{\cellcolor{secondbg}{#1}}

\begin{table*}[t]
\centering
\vspace{-1.0em}
\caption{\textbf{Quantitative evaluation on long video generation.} We report VBench-Long~\citep{vbench} metrics and drift metrics (color drift and motion drift) on 120- and 240-second videos, with methods grouped by base model.}
\vspace{8pt}
\label{tab:main}
\resizebox{\textwidth}{!}{%
\begin{tabular}{l cccccc !{\vrule width 0.8pt} cc}
\toprule
Method & \makecell{Aesthetic\\Quality} $\uparrow$ & \makecell{Background\\Consistency} $\uparrow$ & \makecell{Imaging\\Quality} $\uparrow$ & \makecell{Motion\\Smoothness} $\uparrow$ & \makecell{Subject\\Consistency} $\uparrow$ & \makecell{Dynamic\\Degree} $\uparrow$ & \makecell{Color\\Drift} $\uparrow$ & \makecell{Motion\\Drift} $\uparrow$ \\
\midrule
\multicolumn{9}{c}{\textit{120 seconds}} \\
\midrule
\rowcolor{groupbg}
\multicolumn{9}{l}{\textit{Results on Self-Forcing}} \\
Self-Forcing            & 49.95 & \second{96.17} & 61.60 & \first{\textbf{98.26}} & 96.20 & 30.04 & 25.54 & 46.88 \\
Deep Forcing            & \second{57.26} & \first{\textbf{96.28}} & 65.94 & \second{98.21} & \first{\textbf{97.22}} & 45.49 & \second{59.65} & 71.88 \\
$\infty$-RoPE           & 55.54 & 95.58 & 67.42 & 97.64 & 96.18 & 62.30 & 54.15 & \second{85.16} \\
MemRoPE                 & 55.45 & 95.78 & \second{67.70} & 97.69 & 96.32 & \second{62.90} & 59.25 & 82.81 \\
\textbf{ID-Forcing (Ours)}    &  \first{\textbf{58.23}}        & 96.13 & \first{\textbf{68.20}} & 97.77 & \second{96.76} & \first{\textbf{65.65}} & \first{\textbf{72.37}} & \first{89.06} \\
\addlinespace[2pt]
\rowcolor{groupbg}
\multicolumn{9}{l}{\textit{Results on LongLive}} \\
LongLive               & \first{59.25} & \first{96.52} & 67.32 & \first{98.68} & \second{97.57} & 42.55 & 59.98 & 91.41  \\
Deep Forcing            & 58.31 & \second{96.48} & 66.12 & 98.49 & 97.55 & 45.14 & 60.24 & 89.84  \\
$\infty$-RoPE           & 56.90 & 96.22 & 66.53 & 98.53 & 97.12 & \second{52.81} & 58.20 & 95.31  \\
MemRoPE                 & 57.63 & 96.39 & \second{68.40} & \second{98.65} & 97.30 & 49.18 & \second{65.49} & \first{96.88}  \\
\textbf{ID-Forcing (Ours)}            & \second{58.99} & \second{96.48} & \first{68.73} & 98.57 & \first{97.87} & \first{55.01} & \first{70.94} & \second{96.09}  \\
\midrule
\multicolumn{9}{c}{\textit{240 seconds}} \\
\midrule
\rowcolor{groupbg}
\multicolumn{9}{l}{\textit{Results on Self-Forcing}} \\
Self-Forcing            & 44.90 & \first{96.21} & 58.38 & \first{97.97} & 95.59 & 29.61 & 20.98 & 56.25  \\
Deep Forcing            & \second{56.27} & 95.93 & 65.63 & 97.83 & \second{96.59} & 58.82 & \second{62.52} & \second{88.28}  \\
$\infty$-RoPE           & 55.39 & 95.61 & \second{67.48} & 97.61 & 96.17 & \second{61.98} & 55.49 & 87.50 \\
MemRoPE                  & 54.29 & 95.83 & 67.36 & \second{97.96} & 96.47 & 59.75 & 61.41 & 85.16  \\
\textbf{ID-Forcing (Ours)}              & \first{57.72} & \second{96.10} & \first{67.88} & 97.62 & \first{96.70} & \first{64.76} & \first{70.34} & \first{92.19}  \\
\addlinespace[2pt]
\rowcolor{groupbg}
\multicolumn{9}{l}{\textit{Results on LongLive}} \\
LongLive              & \second{58.94} & \first{\textbf{96.58}} & 67.88 & \first{\textbf{98.74}} & \first{\textbf{97.64}} & 39.79 & 55.64 & 85.16  \\
Deep Forcing            & 57.91 & 96.27 & 66.61 & 98.40 & 97.17 & 46.02 & 57.08 & 89.06  \\
$\infty$-RoPE           & 56.63 & 96.14 & 67.07 & 98.48 & 96.97 & \second{53.95} & 56.72 & 94.53 \\
MemRoPE                 & 57.64 & 96.35 & \second{68.58} & \second{98.66} & 97.30 & 48.76 & \second{64.23} & \first{\textbf{96.88}} \\
\textbf{ID-Forcing (Ours)}            & \first{\textbf{60.44}} & \second{96.40} & \first{\textbf{68.79}} & 98.50 & \second{97.57} & \first{\textbf{56.13}} & \first{\textbf{71.76}} & \second{96.09}  \\
\bottomrule
\end{tabular}%
}
\end{table*}
\subsection{Main Results}
\label{sec:main_results}

\textbf{Quantitative Results.} \Cref{tab:main} reports VBench-Long~\citep{vbench} metrics for 120- and 240-second videos, built on top of two base models. Compared with all baselines, ID-Forcing achieves substantial gains on both drift metrics, indicating that visual quality and motion dynamics remain stable over long durations. This resistance to motion degradation is further shown by Dynamic Degree, where our method achieves the highest score in every setting. In contrast, base models such as Self-Forcing and LongLive exhibit markedly lower dynamic degree, particularly at longer durations, reflecting a tendency toward frame freezing as generation extends.

Notably, our method also performs well on Aesthetic Quality, Imaging Quality, and Subject Consistency. The one exception is Motion Smoothness, which measures local frame-to-frame stability and is therefore inherently biased toward methods with lower dynamic degree; on all other metrics, ID-Forcing remains competitive with or superior to the baselines. This result suggests that our method mitigates long-horizon drift without sacrificing generation quality.

\begin{figure}[t]
    \centering
    \includegraphics[width=1.\linewidth]{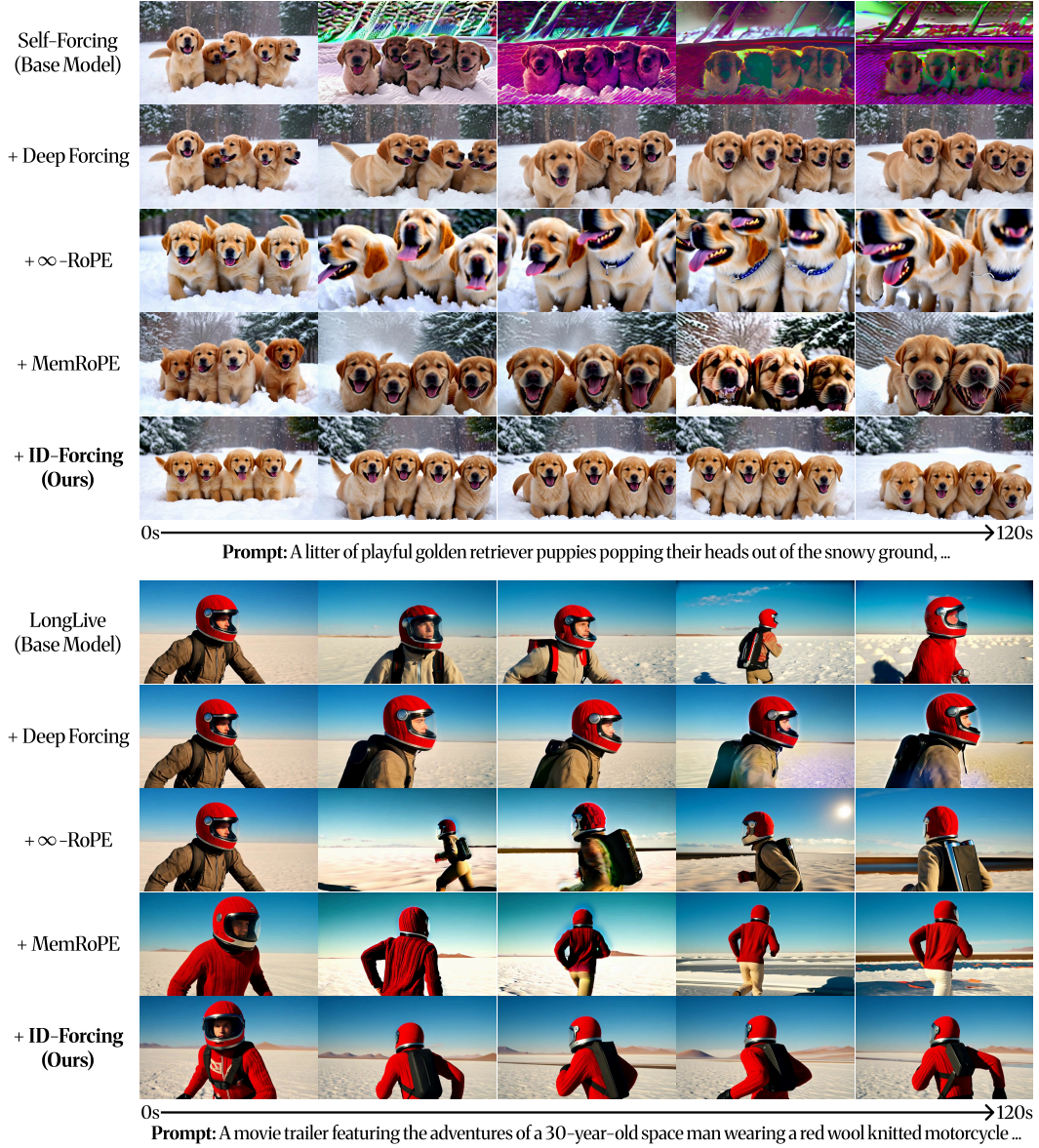}
    \vspace{-.6cm}
    \caption{
        \textbf{Qualitative comparison on 2-minute videos.} 
        Full videos are available at this \href{https://in-distribution-forcing.github.io/}{link}.
    }
    \label{fig:main_qual}
    \vspace{-.2cm}
\end{figure}

\textbf{Qualitative Results.} \Cref{fig:main_qual} shows qualitative comparisons on 2-minute video generation. On the Self-Forcing base model, the base model itself suffers severe color and motion collapse over time. Adding Deep Forcing stabilizes color but subject consistency still degrades, with the puppies' identities gradually drifting. ∞-RoPE exhibits color saturation alongside a similar loss of subject consistency, while MemRoPE shows unnatural color shifts and changing textures across frames. In contrast, ID-Forcing remains stable throughout, preserving both subject consistency and color fidelity.

On the LongLive base model, color drift is generally less severe across all methods, but subject consistency remains a challenge: the base model shows the subject's clothing color changing over time. Deep Forcing suppresses motion and introduces color artifacts, including unnatural brightening and a bluish tint. Both ∞-RoPE and MemRoPE exhibit highly inconsistent motion throughout generation. ID-Forcing, by contrast, maintains subject consistency, sustains motion until the end of the sequence, and preserves accurate color throughout.

\begin{wraptable}{r}{0.5\textwidth}
  \vspace{-2.1em}
  \caption{\textbf{User study.} Preference rate (\%) of ID-Forcing over each baseline in pairwise comparisons (Baseline vs.\ Ours). Each value denotes the percentage of participants who preferred ours ($\uparrow$).}
  \label{tab:user_onecol}
  \vspace{.2cm}
  \centering
  \footnotesize
  \setlength{\tabcolsep}{2.5pt}
  \begin{tabular*}{\linewidth}{l@{\extracolsep{\fill}}*{6}{>{\fontsize{7.8pt}{8.6pt}\selectfont}c}}
  \toprule
  \textbf{Method} &
  {\footnotesize\makecell{Color\\Cons.}} &
  {\footnotesize\makecell{Bg.\\Cons.}} &
  {\footnotesize\makecell{Subj.\\Cons.}} &
  {\footnotesize\makecell{Dyn.\\Deg.}} &
  {\footnotesize\makecell{Temp.\\Flick.}} &
  {\footnotesize\makecell{Over.\\Pref.}} \\
  \midrule
  Self-Forcing  & 100.0 & 99.0 & 96.0 & 93.0 & 98.0 & 100.0 \\
  Deep Forcing  &  90.0 & 84.0 & 85.0 & 81.0 & 97.0 &  95.0 \\
  $\infty$-RoPE &  91.0 & 90.0 & 87.0 & 75.0 & 91.0 &  93.0 \\
  MemRoPE       &  88.0 & 82.0 & 78.0 & 59.0 & 89.0 &  86.0 \\
  \bottomrule
  \end{tabular*}
  \vspace{-1.2em}
\end{wraptable}

\textbf{User Study.} We conducted a user study to complement our quantitative results. Following the 2AFC protocol used in prior works~\citep{deepforcing,memrope}, we extend both the evaluation metrics and the number of prompts to specifically capture long-horizon degradation. 20 participants compared our method against 4 Self-Forcing-based baselines across 20 diverse prompts, each judging a single Baseline-vs-Ours pair per prompt across six dimensions: Color Consistency, Background Consistency, Subject Consistency, Dynamic Degree, Temporal Flickering, and Overall Preference (details in App.~\S\ref{app:user_study}). As shown in \cref{tab:user_onecol}, participants consistently preferred our method over all baselines across every evaluated dimension.

\subsection{Ablation Studies}
\label{sec:ablation}
\begin{figure}[t]
    \centering
    \includegraphics[width=1.\linewidth]{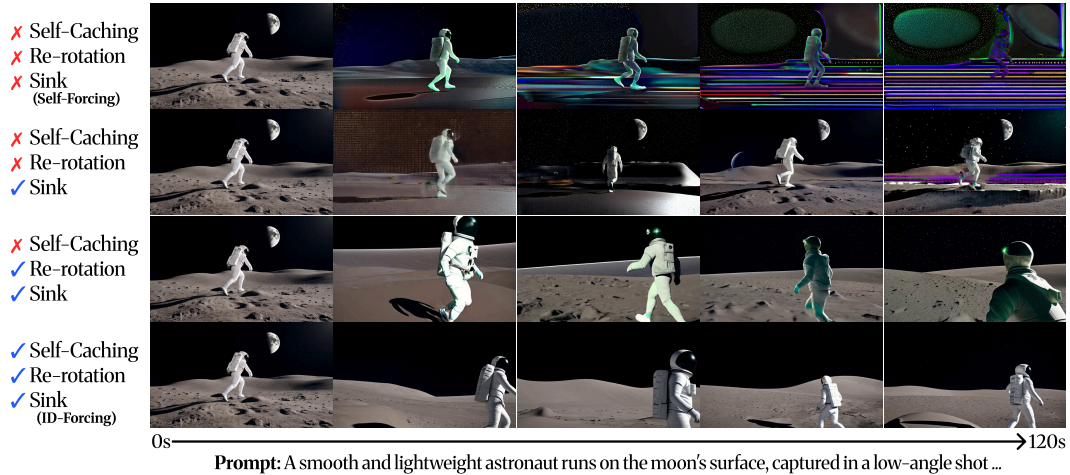}
    \vspace{-.6cm}
    \caption{
       \textbf{Qualitative ablation on 2-minute video.} Components are added from top to bottom. Without self-caching, colors and textures still drift even with the sink and re-rotation, whereas the full ID-Forcing preserves the scene and color throughout the 2-minute video.
    }
    \label{fig:ablation}
    \vspace{-.4cm}
\end{figure}
We ablate the three components of ID-Forcing: self-caching for KV caching, and, for KV conditioning, retaining the first chunk as a sink and re-rotating the window into the trained positional range.
As shown in \cref{fig:ablation}, without any of them, Self-Forcing drastically collapses beyond the training horizon.
Retaining the first chunk largely stabilizes generation, consistent with the known empirical effect of attention sinks, but colors and textures still drift, frames flicker, and the video periodically resets to an earlier scene.
Adding re-rotation improves this further, yet drifting persists, as KV caching remains out-of-distribution.
Self-caching brings KV caching in-distribution as well, greatly reducing color and texture drifting while preserving the dynamics.
Quantitative results are in \cref{tab:ablation} of App.~\S\ref{app:more_results}.

\section{Related Works}
\label{sec:related_works}
\textbf{Video Diffusion Models.}
Video diffusion models with bidirectional attention produce high-quality clips~\citep{sora2, wan, hunyuanvideo, cogvideox}, but attend over the entire sequence at once, so their cost grows quadratically with length and the output length is fixed at sampling time.
Autoregressive (AR) formulations instead factorize the video into chunks generated in sequence, each conditioned on the KV cache of its predecessors~\citep{diffusionforcing, causvid, selfforcing}.
This makes the cost linear, admits streaming output, and in principle allows a video of arbitrary length to be produced at test time.
The AR factorization, however, feeds each chunk back as the condition for the next, so any deviation from the training distribution accumulates along the rollout, a failure known as \emph{drifting}~\citep{diffusionforcing, selfforcing}.

\textbf{Addressing Drifting in AR Video Diffusion.}
The most straightforward way is to train on longer contexts~\citep{selfforcingpp, rollingforcing, longlive, contextforcing,rewardforcing,skyreelsv2,magi}.
However, this does not remove the boundary, since training always covers a fixed horizon while the AR formulation exists to continue past whatever length was trained. Moreover, this is infeasible with extremely long videos since the cost of extending that horizon during training grows with the target length.
A test-time extrapolation method is therefore necessary regardless of how the model was trained, and the two directions are orthogonal: a better-trained backbone still requires a rule for what to do beyond its horizon.

Test-time methods approach this by managing the KV cache during generation.
Deep Forcing~\citep{deepforcing} keeps early KV entries as sinks, realigns their temporal RoPE to the current timeline, and compresses the rest.
$\infty$-RoPE~\citep{infinityrope} reassigns temporal indices so that the offsets seen at inference stay within the trained range.
MemRoPE~\citep{memrope} caches keys without rotation and aggregates evicted entries into memory tokens, applying RoPE online at attention time.
All of these act on \emph{KV conditioning}, leaving untouched how each KV entry was written.
In contrast, we address both \emph{KV caching} and \emph{KV conditioning}, keeping test-time extrapolation within the training distribution.

\section{Conclusion}
\label{sec:conclusion}

In this paper, we introduce \textbf{In-Distribution Forcing}, a training-free framework designed to enable stable long-horizon video generation. By analyzing the autoregressive (AR) model’s key-value (KV) dynamics, we identify the \emph{KV-provenance problem}: a critical out-of-distribution (OOD) failure mode originating at the caching level. To resolve this, we propose a two-level KV management rule: at the caching level, we apply self-caching for KV caching; at the conditioning level, we strictly bound the window length ($L \le N{-}1$) while permanently fixing the first chunk $\kappa_0$. Extensive quantitative and qualitative evaluations demonstrate that our method achieves highly competitive performance across diverse durations, backbones, and metrics. Most importantly, our framework effectively addresses the quality drift issue inherent to autoregressive models, maintaining stable video dynamics even over minute-scale generations.
\clearpage

\section*{AI Use Statement}
We did not use generative AI tools for any tasks with required disclosure under the ICLR 2027 policy. We used generative AI tools only to improve the readability of the manuscript, such as polishing sentences and checking grammar. All AI-assisted edits were reviewed by the authors, and we take full responsibility for the content of this work.
\bibliography{iclr2027_conference}
\bibliographystyle{iclr2027_conference}

\clearpage
\appendix

\section*{\textcolor{black}{Appendix}}
\vspace{-.4cm}

\definecolor{firstbg}{HTML}{a4c1ff}  
\definecolor{secondbg}{HTML}{e0eaff} 
\definecolor{groupbg}{HTML}{F5F5F5}   

\begin{table*}[h]
\centering
\caption{Quantitative evaluation on VBench-Long~\citep{vbench} (30s and 60s).}
\vspace{8pt}
\label{tab:appendix}
\resizebox{\textwidth}{!}{%
\begin{tabular}{l cccccc !{\vrule width 0.8pt} cc}
\toprule
Method & \makecell{Aesthetic\\Quality} $\uparrow$ & \makecell{Background\\Consistency} $\uparrow$ & \makecell{Imaging\\Quality} $\uparrow$ & \makecell{Motion\\Smoothness} $\uparrow$ & \makecell{Subject\\Consistency} $\uparrow$ & \makecell{Dynamic\\Degree} $\uparrow$  & \makecell{Color\\Drift} $\uparrow$ & \makecell{Motion\\Drift} $\uparrow$ \\
\midrule
\multicolumn{9}{c}{\textit{30 seconds}} \\
\midrule
\rowcolor{groupbg}
\multicolumn{9}{l}{\textit{Results on Self-Forcing}} \\
Self-Forcing            & 56.85 & 96.06 & 67.48 & \first{\textbf{98.37}} & 96.33 & 44.58 & 37.45 & 60.94 \\
Deep Forcing            & \second{57.79} & 96.05 & 67.00 & \second{98.13} & \first{\textbf{96.79}} & 59.22 & 63.61 & 82.03 \\
$\infty$-RoPE           & 57.38 & 95.87 & \second{67.84} & 97.80 & 96.65 & 64.79 & 56.77 & 85.94 \\
MemRoPE                 & 57.32 & \second{96.08} & 67.73 & 98.05 & \second{96.78} & \first{\textbf{66.35}} & \second{65.61} & \second{89.84} \\
\textbf{ID-Forcing (Ours)}           & \first{\textbf{58.40}} & \first{\textbf{96.15}} & \first{\textbf{68.20}} & 97.67 & 96.70 & \second{65.94} & \first{\textbf{72.13}} & \first{\textbf{95.31}} \\
\addlinespace[2pt]
\rowcolor{groupbg}
\multicolumn{9}{l}{\textit{Results on LongLive}} \\
LongLive                & \second{59.03} & \second{96.64} & 68.15 & \second{98.72} & 97.75 & 45.10 & 59.65 & 94.53 \\
Deep Forcing            & 58.91 & \first{\textbf{96.80}} & 68.20 & \second{98.72} & \first{\textbf{98.06}} & 43.75 & \second{65.81} & 91.41 \\
$\infty$-RoPE           & 58.41 & 96.54 & 67.78 & 98.66 & 97.57 & \second{48.54} & 59.21 & \first{\textbf{96.88}}  \\
MemRoPE                 & 58.83 & 96.55 & \second{68.74} & 98.69 & 97.56 & 46.98 & 63.61 & 91.41 \\
\textbf{ID-Forcing (Ours)}           & \first{\textbf{60.61}} & 96.60 & \first{\textbf{69.01}} & 98.67 & \second{97.89} & \first{\textbf{49.48}} & \first{\textbf{73.14}} & \second{95.31} \\
\midrule
\multicolumn{9}{c}{\textit{60 seconds}} \\
\midrule
\rowcolor{groupbg}
\multicolumn{9}{l}{\textit{Results on Self-Forcing}} \\
Self-Forcing            & 54.91 & 96.06 & 65.60 & \second{97.93} & 96.33 & 36.82 & 30.61 & 53.12 \\
Deep Forcing           & \second{57.65} & \first{\textbf{96.33}} & 67.03 & \first{\textbf{98.31}} & \first{\textbf{97.21}} & 50.47 & \second{65.07} & 74.22 \\
$\infty$-RoPE           & 56.16 & 95.69 & \second{67.76} & 97.67 & 96.46 & 63.25 & 55.87 & 83.59 \\
MemRoPE                 & 56.45 & 95.98 & 66.64 & 97.88 & 96.68 & \second{63.46} & 63.43 & \second{89.84} \\
\textbf{ID-Forcing (Ours)}           & \first{\textbf{58.69}} & \second{96.15} & \first{\textbf{68.41}} & 97.90 & \second{96.80} & \first{\textbf{65.83}} & \first{\textbf{71.91}} & \first{\textbf{92.97}} \\
\addlinespace[2pt]
\rowcolor{groupbg}
\multicolumn{9}{l}{\textit{Results on LongLive}} \\
LongLive                & \first{\textbf{59.47}} & \second{96.60} & 68.09 & \first{\textbf{98.73}} & 97.60 & 44.11 & 59.69 & 93.75 \\
Deep Forcing            & 58.96 & \first{\textbf{96.76}} & 67.27 & 98.70 & \second{97.85} & 45.39 & \second{62.89} & 92.19 \\
$\infty$-RoPE           & 57.97 & 96.47 & 67.38 & 98.65 & 97.38 & \second{50.44} & 59.77 & \first{\textbf{96.88}} \\
MemRoPE                 & 58.25 & 96.48 & \second{68.22} & \second{98.71} & 97.46 & 47.32 & 60.97 & 92.97 \\
\textbf{ID-Forcing (Ours)}           & \second{59.33} & 96.51 & \first{\textbf{68.95}} & 98.61 & \first{\textbf{97.87}} & \first{\textbf{53.88}} & \first{\textbf{71.76}} & \second{96.09} \\
\bottomrule
\end{tabular}%
}
\end{table*}
\begin{table}[t]
\centering
\caption{Quantitative results of ablation study on VBench-Long~\citep{vbench} (120s).}
\vspace{8pt}
\label{tab:ablation}
\resizebox{\textwidth}{!}{%
\begin{tabular}{l cccccc !{\vrule width 0.8pt} cc}
\toprule
Method & \makecell{Aesthetic\\Quality} $\uparrow$ & \makecell{Background\\Consistency} $\uparrow$ & \makecell{Imaging\\Quality} $\uparrow$ & \makecell{Motion\\Smoothness} $\uparrow$ & \makecell{Subject\\Consistency} $\uparrow$ & \makecell{Dynamic\\Degree} $\uparrow$ & \makecell{Color\\Drift} $\uparrow$ & \makecell{Motion\\Drift} $\uparrow$ \\
\midrule
Self-Forcing                    & 49.95 & \first{96.17} & 61.60 & \first{98.26} & \second{96.20} & 30.04 & 25.54 & 46.88 \\
\quad + Sink                    & \second{55.23} & 95.59 & 64.68 & 96.88 & 95.70 & 46.90 & 48.65 & 67.19 \\
\quad + Re-rotation             & 54.14 & 95.68 & \second{66.89} & 97.36 & 96.08 & \second{65.44} & \second{60.66} & \second{87.50} \\
\quad + Self-Caching \textbf{(ID-Forcing)} & \first{58.23} & \second{96.13} & \first{68.20} & \second{97.77} & \first{96.76} & \first{65.65} & \first{72.37} & \first{89.06} \\
\bottomrule
\end{tabular}%
}
\end{table}
\section{Additional Results}
\label{app:more_results}
\textbf{Quantitative Results.} We provide additional VBench-Long~\citep{vbench} metrics on 30- and 60-second videos to complement the 120- and 240-second results reported in the main paper. Consistent with our main findings, ID-Forcing achieves substantial improvements on both Color Drift and Motion Drift, and likewise attains the highest Dynamic Degree among all compared methods. 

Comparing across all four durations (30, 60, 120, and 240 seconds), several baselines exhibit a clear decline in performance as generation length increases, particularly on Dynamic Degree and the drift metrics (\textit{e.g.} Self-Forcing Dynamic Degree collapses from \textbf{44.58} at 30 seconds to \textbf{29.61} at 240 seconds), reflecting a growing tendency toward frame freezing and quality degradation over longer horizons. In contrast, ID-Forcing maintains largely consistent performance across all durations (\textit{e.g.} \textbf{65.94} at 30 seconds to \textbf{64.76} at 240 seconds), indicating that our method is robust to extrapolation and does not suffer from the same long-horizon degradation observed in the baselines.

\begin{figure}[t]
    \centering
    \includegraphics[width=1.\linewidth]{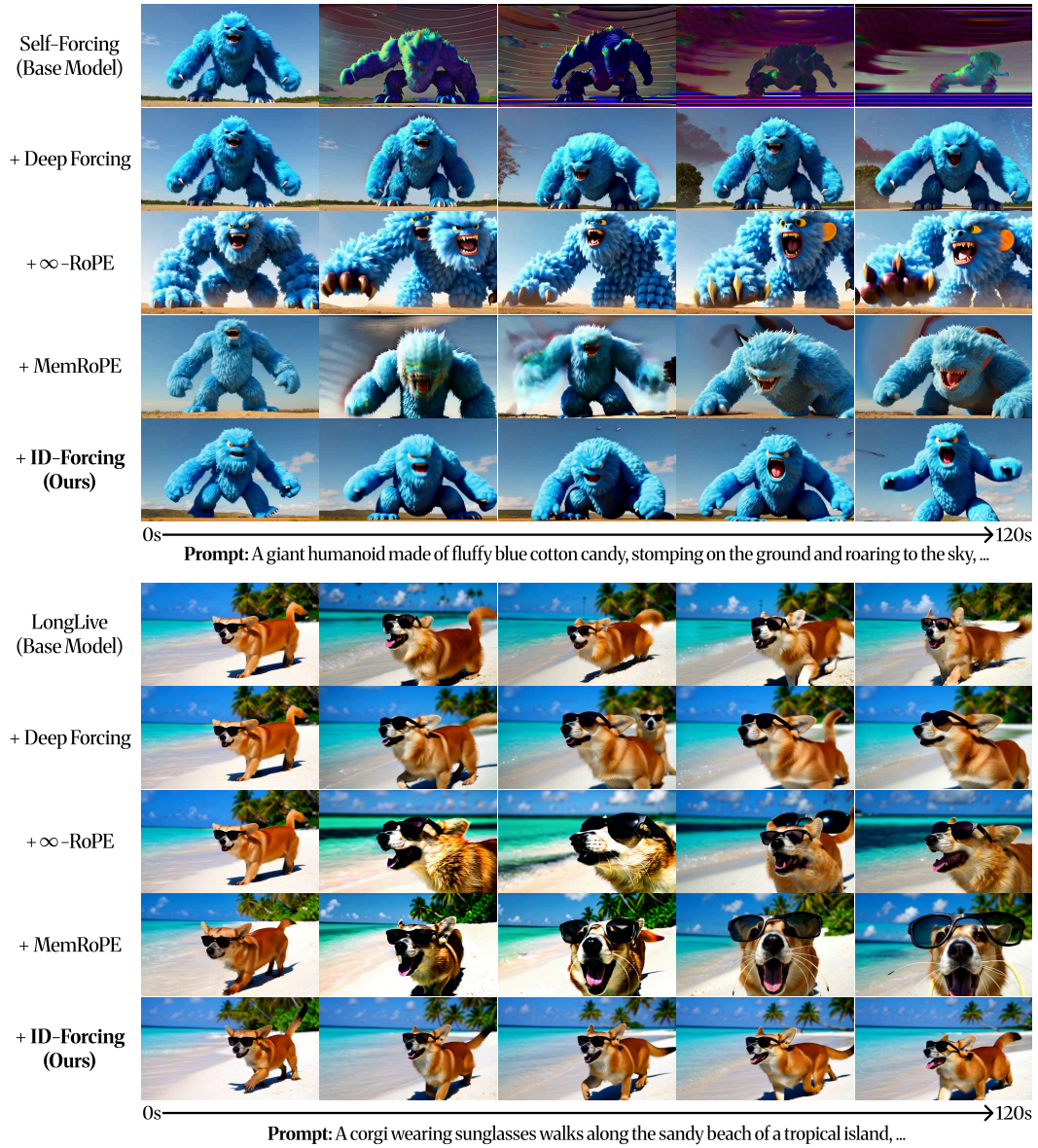}
    \vspace{-.6cm}
    \caption{
        Additional Qualitative results on 2-minute videos.
    }
    \label{fig:add_qual}
    \vspace{-.4cm}
\end{figure}

\textbf{Qualitative Results.} We provide additional qualitative comparisons on 2-minute video generation in \Cref{fig:add_qual}. Additional video results spanning all 4 durations (30s, 60s, 120s, 240s), including our paper reported results, can be found here: \red{\url{https://in-distribution-forcing.github.io/}}.

\textbf{Ablation Results.} \cref{tab:ablation} reports the quantitative evaluations of ablation study, measured on Self-Forcing at 120 seconds with components added cumulatively.

Self-Forcing scores highest on Background Consistency and Motion Smoothness, but this reflects its collapse rather than its quality: once the video degenerates, consecutive frames become nearly identical, which both metrics reward. Its Dynamic Degree of 30.04 and Color Drift of 25.54 show what the frames make visible.

Adding the sink raises Dynamic Degree and Color Drift substantially, confirming that generation is stabilized, yet Color Drift remains at 48.65 and Motion Drift at 67.19, consistent with the drifting and periodic resets still visible in \cref{fig:ablation}. Re-rotation improves Dynamic Degree and both drift metrics further, but Color Drift stays at 60.66, since KV caching is still out of distribution.

With self-caching, ID-Forcing yields the best score on every metric except Background Consistency and Motion Smoothness, which favor the collapsed output of Self-Forcing. Color Drift improves from 60.66 to 72.37 and Aesthetic Quality from 54.14 to 58.23, matching the reduced drifting and preserved dynamics observed in the frames.

\section{Metric Description}
\label{app:impl_detail}

\textbf{Color Drift (HSV Histogram).} Following~\citet{ttc}, we quantify changes in color distribution over the course of a generated sequence by comparing the color histograms of the first and last frames. Given the initial frame and final frame of a generated video, we convert both to HSV space and compute an $L_1$-normalized histogram of the Hue channel using 180 bins, denoted $h_{\text{start}}, h_{\text{end}} \in \mathbb{R}^{180}$, with $\|h_{\text{start}}\|_1 = \|h_{\text{end}}\|_1 = 1$. We measure the $L_1$ distance $\|h_{\text{start}} - h_{\text{end}}\|_1$ between the two histograms, which lies in $[0, 2]$.

\textbf{Motion Drift (Dynamic Degree Difference).} To measure how well motion is sustained throughout a generated sequence, we compute the Dynamic Degree~\citep{vbench} separately over the first 5 seconds and the last 5 seconds of the video, denoted $d_{\text{start}}$ and $d_{\text{end}}$, respectively. We measure the difference $|d_{\text{start}} - d_{\text{end}}|$, which lies in $[0, 1]$ and captures the extent to which motion established early in the video is preserved or lost by the end of generation. A large difference reflects degradation, most commonly a collapse toward static, near-frozen frames.

\textbf{Score Calibration.} For consistency with VBench-Long metrics, where higher is better and scores range from 0 to 100, we convert both drift measures into scores by normalizing each distance by its maximum and subtracting it from 1:
\begin{equation}
    \text{Color Drift} = 100 \left(1 - \tfrac{1}{2}\|h_{\text{start}} - h_{\text{end}}\|_1\right), \qquad
    \text{Motion Drift} = 100 \left(1 - |d_{\text{start}} - d_{\text{end}}|\right).
\end{equation}
Both scores are computed for each generated video and averaged over all 128 prompts.
A higher score thus indicates that color and motion dynamics remain more stable throughout generation. All drift results in the main paper are reported in this calibrated form.

\section{User Study Details}
\label{app:user_study}

\begin{figure}[t]
    \centering
    \includegraphics[width=1.\linewidth]{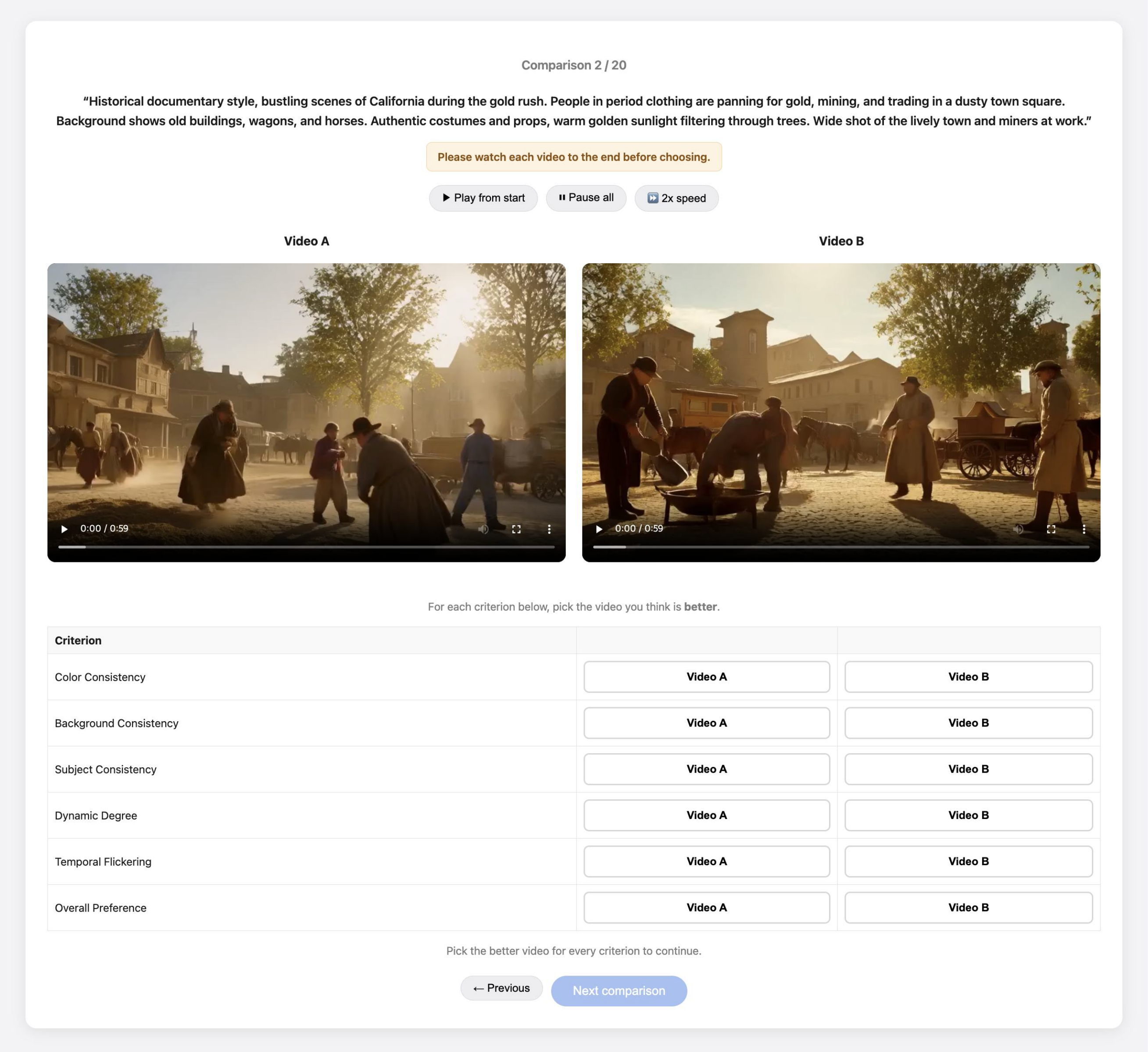}
    \vspace{-.4cm}
    \caption{
        User study interface. 
    }
    \label{fig:user_study_interface}
    \vspace{-.4cm}
\end{figure}

\begin{figure}[h]
    \centering
    \includegraphics[width=1.\linewidth]{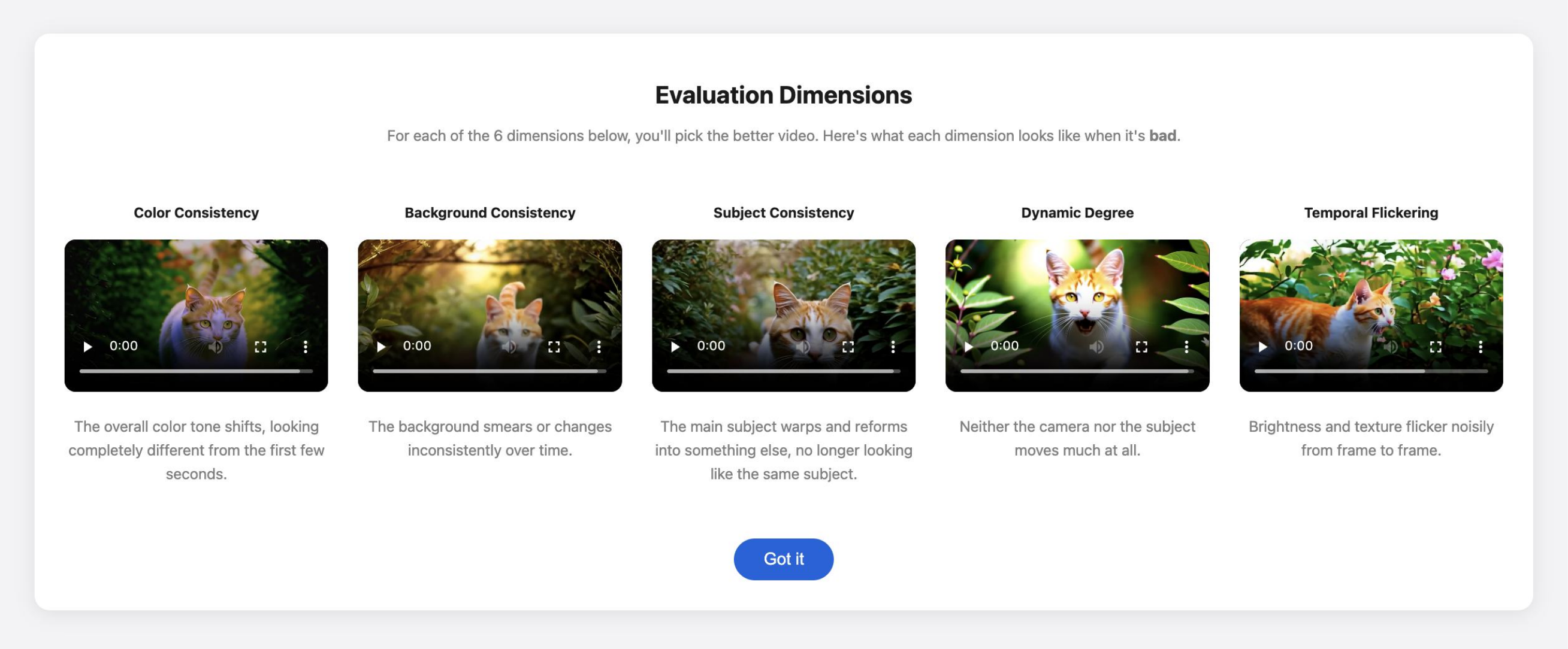}
    \vspace{-.4cm}
    \caption{
        Criteria explanation.
    }
    \label{fig:user_study_explanation}
    \vspace{-.4cm}
\end{figure}

For each of the 20 prompts, participants were shown two anonymized videos side by side: one generated by our method and one by a randomly assigned baseline. They were asked to indicate their preference along Color Consistency, Background Consistency, Subject Consistency, Dynamic Degree, Temporal Flickering, and Overall Preference. \cref{fig:user_study_interface} demonstrates our user study interface and \cref{fig:user_study_explanation} show our criteria explanation page.

To ensure fair comparisons across all baselines, we employed a balanced Latin-square design for baseline assignment. We divided the 20 prompts into 5 groups of 4 and split participants into 4 groups. Within each group of 4 prompts, we rotated the baseline assignment so that each of the 4 baselines appeared exactly once. As a result, every participant saw each baseline exactly 5 times (5 groups × 1 time each), and every prompt was compared against all 4 baselines equally across the full set of participants. This ensured that no baseline was unfairly matched with easier or harder prompts, keeping all comparisons balanced.

\end{document}